\pdfoutput=1
\documentclass[11pt]{article}

\usepackage[final]{acl}
\usepackage{tgtermes}
\usepackage{latexsym}
\usepackage[T5]{fontenc}   % Vietnamese diacritics (Điều, phải, ...)
\usepackage[utf8]{inputenc}
\usepackage{microtype}
\microtypesetup{protrusion=false}
\usepackage{graphicx}
\usepackage{amsmath}
\usepackage{amssymb}
\usepackage{booktabs}
\usepackage{array}
\usepackage{multirow}
\usepackage{enumitem}
\usepackage{placeins}
\usepackage{stfloats}
\title{Cross-Lingual Legal QA for Vietnamese Labour Law: Retrieval, Translation, and Verifier-Guided Correction}

\author{
\textbf{Nguyen Minh Chi}\textsuperscript{1},
\textbf{Mo El-Haj}\textsuperscript{1,3},
\textbf{Nguyen Ha Thanh}\textsuperscript{1},
\textbf{Dawn Knight}\textsuperscript{2} \and
\textbf{Paul Rayson}\textsuperscript{3,1} \\
\textsuperscript{1}VinUniversity, Vietnam \quad
\textsuperscript{2}Cardiff University, UK \quad
\textsuperscript{3}Lancaster University, UK
}
\date{}

\begin{document}

\maketitle

\begin{abstract}
Cross-lingual legal question answering must retrieve statutes across
languages while preventing unsupported legal claims.
We introduce a bilingual evaluation suite of 231 Vietnamese--English
question--answer pairs from Vietnamese labour law. Of these, 75 are
additionally annotated for five challenging legal reasoning phenomena.
We evaluate a verifier-guided pipeline that decomposes answers into claims,
checks citation reachability and entailment, and corrects citation failures
and contradictions.
We also introduce six automatic diagnostics for faithfulness to retrieved
evidence, covering citations, modality, exceptions, procedures, conclusions,
and evidential support.
Experiments show that learned-sparse retrieval performs poorly for
English-to-Vietnamese retrieval (R@5~=~0.032), whereas dense retrieval
reaches 0.358 and slightly outperforms hybrid retrieval.
Translation placement has no statistically detectable effect on these
automatic diagnostics in our controlled comparison and supporting
sensitivity analyses.
Verifier-guided correction improves citation preservation by
$0.022$--$0.034$ at the system level but produces no reliable gains in the remaining dimensions.
Human evaluation further shows that the automatic diagnostics do not fully align with human judgements of answer quality.
\end{abstract}

% ============================================================
\section{Introduction}
% ============================================================

Legal question answering (QA) systems are increasingly studied and piloted to help
users navigate complex statutory frameworks, yet most existing systems
operate within a single language.
For legal domains in the Global South, statutes are often available only
in the national language, while a substantial fraction of users, including
foreign investors, international organisations, and cross-border legal
practitioners, require access in English.
Vietnamese labour law is a prototypical case: the Labour Code
(\textit{Bộ Luật Lao Động}) governs employment relationships in a large,
internationally connected labour market, yet cross-lingual QA systems for
this corpus remain largely unstudied in the NLP literature.

Cross-lingual legal QA introduces two additional sources of failure.
First, retrieval degrades when English queries are matched against Vietnamese statutory text with little
effective token overlap, making lexical matching difficult
and motivating dense multilingual retrieval.
Second, the generator operates over cross-lingual
context, either a translated query or translated statutory evidence,
and may produce answers that are fluent and plausible but unsupported by the retrieved statutory evidence,
for example by citing the wrong provision, ignoring exception clauses,
or inverting modal distinctions between obligation (\textit{shall})
and discretion (\textit{may}).

This paper targets a narrow, defensible claim: cross-lingual legal QA
for Vietnamese statutes is a retrieval, verification, and
translation-faithfulness problem.
We evaluate our approach through automatic diagnostics on all 231 questions
and human judgements on a 27-question sample. The two evaluation views diverge,
motivating their separate analysis.
Recent legal QA work has shown that RAG improves factual grounding in
domain-specific settings \citep{Bea24,Red24,Gok25b}. Nearby multilingual and
low-resource studies focus on retrieval or RAG design rather than answer-level
legal correction \citep{Nguyen25CLIR,Kab25,Aka25b}.
At the same time, recent Vietnamese work provides the missing building
blocks: ViLegalNLI \citep{Duo26} supplies a legal entailment corpus for
claim verification, VLegal-Bench \citep{Don25} supplies hard reasoning cases,
and \citet{Le26} characterise the reasoning errors LLMs produce on Vietnamese
legal text, with incorrect examples and misinterpretation among the most
prevalent categories.
What remains missing is a cross-lingual pipeline that combines these components
with systematic retrieved-evidence diagnostics.

We present three contributions:
\begin{enumerate}[noitemsep,leftmargin=*]
    \item \textbf{A cross-lingual evaluation suite} for Vietnamese labour law QA,
          curated from VLegal-Bench and ALQAC, comprising 231 bilingual
          (VI/EN) question-answer pairs with 75 hard-case annotations
          across five reasoning phenomena, paired with a 953-article
          retrieval corpus that includes post-2024 amending and implementing
          instruments.
    \item \textbf{A verifier-guided correction pipeline} with a two-stage
          verifier (rule-based citation gate + GPT-4o-mini NLI entailment)
          that triggers single-pass correction on citation failures and
          entailment contradictions.
    \item \textbf{Six automatic retrieved-evidence diagnostics} capturing
          citation preservation, modality, exception preservation,
          procedural completeness, conclusion consistency, and support
          preservation, dimensions that generic metrics miss.
\end{enumerate}

We evaluate four cross-lingual configurations and a retrieval mode
ablation, finding that dense retrieval performs best for direct English
queries, that the placement of translation within the pipeline does not measurably change the automatic
diagnostic scores once retrieval is held fixed, and that procedural
completeness receives the lowest diagnostic score across B2--B4.
Together with the retrieval gap, this null result suggests that retrieval may be a
more consequential bottleneck than translation placement in our setting.

% ============================================================
\section{Related Work}
% ============================================================

\subsection{Legal QA with Retrieval-Augmented Generation}

Legal information extraction and question answering form part of a broader
landscape of legal NLP tasks, where challenges such as domain-specific
language, complex document structure, limited annotated data, and the need
for reliable information extraction remain significant
\citep{premasiri2025survey}.
Within this broader context, the application of RAG to legal domains has
produced several domain-specific QA systems.
\citet{Bea24} apply RAG to Quebec automobile insurance law, demonstrating
that passage retrieval substantially improves factual grounding over
ungrounded generation.
\citet{Red24} address attribution for precondition-heavy Dutch statutes,
where answers must cite specific enabling conditions.
\citet{Gok25b} propose learning-to-rank for multi-passage regulatory RAG,
improving the coherence of retrieved context across related statutory provisions.
These systems ground answers in retrieved text but offer no post-generation
verification of whether the answer faithfully reflects what the statute says.

\subsection{Multilingual and Cross-Lingual Legal NLP}

The development of multilingual legal NLP is also supported by increasingly
available cross-lingual legal resources. \citet{el2024multilingual} introduce
the Multilingual Corpus of World's Constitutions (MCWC), providing
constitutional texts across multiple languages and supporting comparative
and multilingual legal NLP research.
\citet{Nguyen25CLIR} introduce Vietnamese--English cross-lingual retrieval
datasets for general and legal domains and improve retrieval through
auxiliary-loss and symmetric training objectives.
\citet{Kab25} develop bilingual QA over English--Bangla police gazettes using
dense multilingual retrieval with LLM-based relevance checking and query
refinement.
\citet{Aka25b} introduce NitiBench and compare parametric, RAG, and
long-context approaches to Thai legal QA, finding that hierarchy-aware RAG
improves coverage and contradiction scores.
\citet{Aka25d} align LLMs for Thai legal QA using semantic similarity rewards,
establishing that task-specific alignment helps even without large legal
training sets.
None of these systems incorporate a dedicated verification step that checks
generated answers against retrieved statutory text.

\subsection{Vietnamese Legal NLP}

\citet{Don25} introduce VLegal-Bench, a cognitively grounded benchmark for
Vietnamese legal reasoning covering multi-step inference, numeric thresholds,
and exception chains.
\citet{Le26} conduct a large-scale evaluation of LLMs on Vietnamese legal
text with an expert-annotated error typology, identifying incorrect examples
and misinterpretation as the most prevalent error categories.
\citet{Duo26} develop ViLegalNLI, a natural language inference dataset for
Vietnamese legal entailment, which motivates the NLI formulation of our
Stage-2 verifier.
Prior Vietnamese--English work addresses cross-lingual legal retrieval
\citep{Nguyen25CLIR}; we study end-to-end answer generation and
post-generation correction over Vietnamese statutes.

\subsection{Legal Faithfulness and Evaluation}

\citet{Gui25} evaluate LLM-generated legal explanations for regulatory
compliance, showing that high classification accuracy does not ensure
trustworthy explanations because models may cite irrelevant or fabricated
provisions.
\citet{Le26} categorise LLM reasoning errors from expert annotation,
including omissions of core answer elements and applicable exception clauses.
These categories motivate our exception and completeness dimensions; our
modality and procedural dimensions are task-specific extensions for
cross-lingual statutory QA.

% ============================================================
\section{Evaluation Suite}
\label{sec:dataset}
% ============================================================

\subsection{Dataset Construction}

We curate a bilingual evaluation suite from two existing Vietnamese legal
QA resources: VLegal-Bench~\citep{Don25} and the 2021 Automated Legal Question
Answering Competition (ALQAC)~\citep{Thanh22}, filtering for questions
grounded in the Vietnamese Labour Code. We also screened a separate
community-derived VietLegal collection, but excluded it from the final suite
because its citation coverage was incomplete.
The final suite contains 231 question-answer pairs from the
Labour Code (Bộ Luật Lao Động, No.~45/2019/QH14) and five implementing
decrees.
Each record includes a Vietnamese question (\textit{question\textsubscript{VI}}),
an English translation (\textit{question\textsubscript{EN}}),
a Vietnamese reference answer, and one or more gold citation articles.
Eighty-eight records (38\%) require citing three or more gold articles,
testing multi-article retrieval recall.

\paragraph{English translation.}
Questions and answers are translated to English using GPT-4o-mini with
a legal-fidelity prompt that explicitly preserves modal distinctions
(\textit{phải} $\to$ \textit{shall},
\textit{không được} $\to$ \textit{must not},
\textit{có thể} $\to$ \textit{may}),
exception clauses (\textit{ngoại trừ} $\to$ \textit{except for}),
and statutory article identifiers verbatim.

\subsection{Hard-Case Annotation}

A total of 75 hard-case questions (32.5\% of the corpus) were annotated across
five reasoning phenomena that stress retrieval and generation. We define a
hard-case question as a QA instance that requires reasoning beyond the direct
retrieval of a single self-contained provision, because the answer depends on
exceptions, cross-references, numeric or temporal conditions, or the synthesis
of multiple provisions. These categories are motivated by failure mode
analyses in \citet{Don25} and \citet{Le26} and cover the main structural
challenges in Vietnamese labour law: exception-conditional rules (the most
frequent, 41 cases), provisions that depend on cross-article definitions (12),
numeric thresholds (11), temporal thresholds (6), and answers that require
synthesising multiple non-adjacent articles (5). The 75 hard-case questions
are a subset of the 231 QA pairs and carry no additional annotation burden
beyond the phenomenon tag.

\subsection{Retrieval Corpus}

The statutory retrieval corpus comprises \textbf{953 statutory articles}
drawn from the Labour Code and its implementing decrees. The articles are
indexed in a Qdrant vector store using BGE-M3 embeddings \citep{Che24}
with 1,024 dimensions. Each article is treated as a single retrieval
chunk, preserving complete statutory units and enabling article-level
citation evaluation.

% ============================================================
\section{Method}
\label{sec:method}
% ============================================================

\subsection{Pipeline Overview}

Our system implements a four-stage pipeline:
\textit{retrieve} $\to$ \textit{generate} $\to$ \textit{verify} $\to$ \textit{correct}.
Table~\ref{tab:components} (Appendix~\ref{app:repro}) summarises each
component, its model, and its role.
The pipeline is modular: retrieval mode and evidence language are
configurable independently of the verifier and evaluator,
enabling the cross-lingual ablation described in Section~\ref{sec:experiments}.

\subsection{Retrieval}

Given an input question, we retrieve the top-$k$ statute articles from
the Vietnamese corpus.
For the monolingual baseline (B1), the Vietnamese question is passed directly.
For query-translation cross-lingual retrieval (B2), GPT-4o-mini translates the
English question to Vietnamese before retrieval.
For direct English-query retrieval (B3, B4), the English question is submitted
to BGE-M3 dense retrieval without translation.
Main pipeline runs use $k=5$; the standalone retrieval ablation uses $k=10$.

\paragraph{Gold-citation robustness.}
Gold labels contain bare article numbers but not instrument identifiers.
We therefore use number-level matching as the primary rule and report a
Labour-Code-only sensitivity analysis in Section~\ref{sec:results} and
Appendix~\ref{app:extra}. Number-level recall should be read as an upper bound.

\subsection{Generation}

Retrieved statute passages and the input question are passed to
Qwen3-8B \citep{Yan25} with a citation-grounded prompt that instructs the model
to reference every factual claim with an explicit article identifier.
For B4, retrieved Vietnamese passages are translated to English by
GPT-4o-mini before the generation prompt is assembled, so the generator
operates entirely in English on English-language statutory text.

\subsection{Two-Stage Verification}
\label{subsec:verifier}

The generated answer is first decomposed into atomic claims
by GPT-4o-mini.
Each claim is tagged with the article numbers it cites and passed through
two sequential verification stages.

\paragraph{Stage 1: Citation Gate (rule-based).}
The gate checks two conditions:
\begin{enumerate}[noitemsep, label=(\alph*)]
    \item \textbf{Citation reachability.} The cited article was present in the retrieved top-$k$.
    \item \textbf{Deontic-signal coverage.} The claim does not ignore
          obligation (\textit{phải}, \textit{không được}) or exception
          (\textit{ngoại trừ}) signals present in the cited passage.
\end{enumerate}
Condition (a) failures are labelled \texttt{FAIL\_CITATION};
condition (b) failures are labelled \texttt{FAIL\_CONDITION}.
\texttt{FAIL\_CONDITION} claims are passed to the final answer
\textit{without correction}: rewriting a claim that missed an exception
without a directly supporting passage risks substituting one distortion
for another.

\paragraph{Stage 2: NLI Entailment (GPT-4o-mini).}
Claims passing Stage 1 are classified as \textsc{supported} (entailed by
the cited passage), \textsc{contradicted} (conflicts with the cited passage),
or \textsc{not\_addressed} (cited passage does not speak to the claim).
Only \textsc{contradicted} claims trigger correction;
\textsc{not\_addressed} claims pass through uncorrected.

\subsection{Correction}

\texttt{FAIL\_CITATION} and \textsc{contradicted} claims are sent to
Qwen3-8B for a single-pass rewrite.
The corrector receives the original claim, the retrieved passage, and the
failure type, and is instructed to revise the claim to be faithful to
the passage while preserving the answer's overall structure.

\subsection{Evaluation Metrics}
\label{subsec:metrics}

\paragraph{Retrieval metrics.}
We report Recall@$k$ ($k \in \{1,3,5,10\}$), Mean Reciprocal Rank Multi-MRR (mean reciprocal rank across gold citation numbers per question,
then averaged over questions),
MAP@10, nDCG@10, and Citation Hit Rate
(CHR: fraction of questions for which at least one gold article
appears in the retrieved top-$k$).

\paragraph{Retrieved-evidence diagnostics (6 dimensions).}
Each final answer is scored on six dimensions:
\textit{citation preservation} (Cit.): statutory identifiers retained;
\textit{modality preservation} (Mod.):  \textit{shall}/\textit{may}/\textit{must not} distinctions correct;
\textit{exception preservation} (Exc.) :  ``except for'' clauses and
numeric/conditional thresholds stated explicitly;
\textit{procedural completeness} (Proc.):  filing deadlines, competent
authorities, and prerequisite steps retained;
\textit{conclusion consistency} (Con.):  legal conclusion and scope
match the source statute;
\textit{support preservation} (Sup.):  claims supported by the cited passage.

Citation, modality, exception, and procedural dimensions are scored by
rule-based keyword and deontic-signal matchers against retrieved passages.
Conclusion consistency and support preservation are scored by GPT-4o-mini
acting as an LLM judge.
The \textbf{overall diagnostic score} is the mean of all applicable sub-scores.
Modality, exception, and procedural sub-scores are not computed for B1
because the automatic matchers were calibrated for English-language outputs;
B1's overall diagnostic score therefore covers only Cit., Con., and Sup.\ and is not
directly comparable to B2--B4.

The automatic scores are an internal diagnostic signal, not expert
validation: four dimensions use rule-based matchers, and the two LLM-judge
dimensions use GPT-4o-mini, which also performs translation, claim
splitting, and verification in this pipeline.
The completed human evaluation in Section~\ref{subsec:human_eval} does not
validate the automatic scores; conclusions about legal correctness therefore
remain provisional.

% ============================================================
\section{Experiments}
\label{sec:experiments}
% ============================================================

\subsection{Baseline Configurations}

Table~\ref{tab:baselines} defines four primary systems (B1--B4) spanning
the cross-lingual design space and two auxiliary controls (C3--C4).
B1 operates monolingually in Vietnamese.
B2 bridges the language gap at the query layer by translating the English
question to Vietnamese before retrieval.
B3 submits the English question directly to dense retrieval without
translation, receiving Vietnamese statute passages for generation.
B4 extends B3 by translating retrieved Vietnamese passages to English
before the generation prompt, so the generator operates entirely in English.
All four baselines share the same retrieval corpus (953 articles),
generator (Qwen3-8B), two-stage verifier, and evaluator;
the only variables are query language, retrieval mode, and evidence language.
C3 and C4 complete the $2{\times}2$ translation design under dense retrieval,
but we treat them as sensitivity controls because their Vietnamese query
translations were generated independently rather than held fixed.

\begin{table*}[t]
\centering
\small
\begin{tabular}{llllll}
\toprule
\textbf{System} & \textbf{Query lang.} & \textbf{Retrieval} & \textbf{Evidence lang.} & \textbf{Answer lang.} & \textbf{Role} \\
\midrule
\multicolumn{6}{l}{\textit{Primary systems}} \\
B1 & Vietnamese          & Hybrid (RRF) & Vietnamese    & Vietnamese & Monolingual reference \\
B2 & EN$\to$VI$^\dagger$ & Hybrid (RRF) & Vietnamese    & English & End-to-end cross-lingual \\
B3 & English             & Dense        & Vietnamese    & English & $2{\times}2$: {--}query, {--}evidence \\
B4 & English             & Dense        & EN$^\ddagger$ & English & $2{\times}2$: {--}query, +evidence \\
\midrule
\multicolumn{6}{l}{\textit{Auxiliary sensitivity controls}} \\
C3 & EN$\to$VI$^\dagger$ & Dense        & Vietnamese    & English & $2{\times}2$: +query, {--}evidence \\
C4 & EN$\to$VI$^\dagger$ & Dense        & EN$^\ddagger$ & English & $2{\times}2$: +query, +evidence \\
\bottomrule
\end{tabular}
\caption{Primary system configurations and auxiliary translation controls
($k=5$, 231 questions).
B1--B4 are the primary systems. B3/B4/C3/C4 form a $2{\times}2$ over query
and evidence translation under dense retrieval; C3/C4 are sensitivity
controls because their query translations were generated independently.
B1 and B2 are end-to-end references outside this controlled comparison.
$^\dagger$Query translated EN$\to$VI before retrieval (GPT-4o-mini).
$^\ddagger$Evidence translated VI$\to$EN before generation (GPT-4o-mini).}
\label{tab:baselines}
\end{table*}

\subsection{Retrieval Mode Ablation}

To isolate the contribution of each retrieval mode independently of
the end-to-end pipeline, we run a standalone retrieval evaluation on
all 231 English questions against the Vietnamese corpus at $k=10$,
comparing dense BGE-M3, hybrid RRF (dense + BGE-M3 sparse), and
BGE-M3 sparse-only retrieval.
This ablation targets the retrieval step shared by B3 and B4.

\subsection{Query-Language vs.\ Retriever Disentanglement}
\label{subsec:disentangle_setup}

B1/B2 use hybrid retrieval while B3/B4 use dense, which confounds
query-language effects with retriever-type effects.
To separate them, we run two additional retrieval controls at $k=10$:
(i) Vietnamese queries with dense retrieval, and
(ii) EN$\to$VI queries with dense retrieval (B2 with dense mode).
Together with the B3-dense (English) and B1/B2-hybrid baselines, these
form a $2{\times}2$ query-language $\times$ retriever-type matrix.

% ============================================================
\section{Results}
\label{sec:results}
% ============================================================

\subsection{Retrieval Performance}

Table~\ref{tab:retrieval} reports retrieval metrics.
B1 achieves the strongest retrieval performance, while query translation
(B2) tracks it closely. Direct English-query systems B3 and B4 trail B1/B2
and share identical retrieval metrics because evidence translation occurs
after retrieval.
Their diagnostic contrast is computed on the
231 questions with outputs from both systems and therefore isolates the
post-retrieval evidence-language condition.

\paragraph{Gold-citation matching.}
Gold citations are recorded as bare article numbers without an identifying
instrument.
Because the corpus spans 20 instruments, a bare number is ambiguous: the
average gold number occurs in 6.1 distinct instruments, and only 3 of 572
are unique to a single one.
We therefore credit a retrieved article when its number matches the gold
number, irrespective of instrument. This permissive rule that over-credits.
Under a strict rule (crediting only the Labour Code itself), R@5 falls by
0.009--0.018 across all six systems; on B3, 10.1\% of top-5 hits are
credited solely through a non-Code instrument.
The ranking of systems is identical under both rules; absolute values should
be read as an upper bound (per-system strict values are reported in
Appendix~\ref{app:extra}).

\begin{table*}[t]
\centering
\small
\begin{tabular}{lrrrrrrrrr}
\toprule
\textbf{Sys.} & \textbf{R@1} & \textbf{R@3} & \textbf{R@5} & \textbf{R@10}
              & \textbf{MRR} & \textbf{MMRR}
              & \textbf{MAP} & \textbf{nDCG} & \textbf{CHR} \\
\midrule
\multicolumn{10}{l}{\textit{(a) Pipeline systems ($k=5$)}} \\
B1 & \textbf{.281} & \textbf{.353} & \textbf{.396} & ---
   & \textbf{.486} & \textbf{.331} & --- & --- & \textbf{.687} \\
B2 & .263 & .335 & .394 & ---
   & .471 & .317 & --- & --- & .662 \\
B3 & .246 & .310 & .358 & \textbf{.456}
   & .453 & .294 & .307 & .379 & .662 \\
B4 & .246 & .310 & .358 & \textbf{.456}
   & .453 & .294 & .307 & .379 & .662 \\
\midrule
\multicolumn{10}{l}{\textit{(b) Retrieval mode --- standalone English queries ($k=10$)}} \\
Dense  & --- & --- & \textbf{.358} & \textbf{.456}
       & \textbf{.453} & \textbf{.295} & \textbf{.307} & \textbf{.380} & \textbf{.662} \\
Hybrid & --- & --- & .350 & .433 & .443 & .287 & .296 & .365 & .636 \\
Sparse & --- & --- & .032 & .060 & .046 & .020 & .020 & .035 & .113 \\
\bottomrule
\end{tabular}
\caption{Retrieval performance over all 231 questions.
R@$k$ = Recall@$k$; MMRR = Multi-MRR; MAP = MAP@10; nDCG = nDCG@10;
CHR = Citation Hit Rate.
Panel (a): end-to-end pipeline runs; Pipeline systems used $k=5$, so R@10/MAP/nDCG are unavailable.
B3 and B4 have identical retrieval results because evidence translation is
post-retrieval.
Panel (b): standalone retrieval ablation over the three modes;
Sparse = BGE-M3 learned sparse vectors.
Bold = best within each panel.}
\label{tab:retrieval}
\end{table*}

\subsection{Retrieval Mode Ablation}

Table~\ref{tab:retrieval}(b) shows the retrieval-mode ablation.
BGE-M3 sparse performs poorly for English-to-Vietnamese retrieval,
while hybrid RRF underperforms dense at every metric. Thus, rank fusion
provides no improvement over dense BGE-M3 in this setting.

Query-language and retriever-type effects are separable.
EN$\to$VI queries achieve R@5~=~0.394 vs.\ 0.358 for direct English queries
under dense retrieval --- a gap present with both retrievers.
Conversely, switching from hybrid to dense \textit{within} the same
query language produces no consistent gain.
The retrieval advantage of translated queries is therefore associated with query language rather than the fusion strategy.

\subsection{Retrieved-Evidence Diagnostics}

Table~\ref{tab:faithfulness} reports the six diagnostics.
Among cross-lingual baselines, \textbf{B4} (English queries + translated
evidence) obtains the highest overall score, followed by \textbf{B3} and
\textbf{B2}. B4 leads on exception preservation, procedural completeness,
and conclusion consistency; B3 leads on citation and support preservation.
Because B3 and B4 retrieve identical passages for all 231 questions
(evidence translation is strictly post-retrieval), the B3--B4 contrast
isolates evidence language.
That contrast is $+0.012$ overall with a 95\% bootstrap confidence
interval of $[-0.002, +0.027]$: directionally favourable to translation,
but not distinguishable from zero.
The remaining three contrasts of the $2{\times}2$ (C3--B3, C4--B4,
C4--C3) point the same way, at $+0.008$ to $+0.013$ with every interval
spanning zero.
C3 and C4 returned identical top-5 passages for only 179 of 231 questions
because their query translations were generated independently.
On this matched subset, C4--C3 remains null at $+0.009$
[$-$0.009, $+$0.028]; we therefore use C3/C4 only as a sensitivity analysis.
\textbf{B2} obtains a lower mean than B4 on every dimension, but B2
differs from B4 in query language, retrieval mode, and evidence language
at once, so this end-to-end gap does not isolate a single factor.

B1's overall diagnostic score is not comparable to cross-lingual baselines:
modality, exception, and procedural sub-scores were not computed for B1
because the automatic matchers were calibrated for English-language outputs,
so B1's overall covers only Cit., Con., and Sup.
Within those three dimensions, B1 achieves the highest conclusion consistency
and support preservation.

\begin{table*}[t]
\centering
\small
\begin{tabular}{lrrrrrrr}
\toprule
\textbf{Sys.} & \textbf{Cit.} & \textbf{Mod.} & \textbf{Exc.}
              & \textbf{Proc.} & \textbf{Con.} & \textbf{Sup.} & \textbf{Ovr.} \\
\midrule
B1 & .405 & ---  & ---  & ---  & .862 & .930 & .732$^\dagger$ \\
B2 & .435 & \textbf{.459} & .593 & .270 & .828 & .877 & .577 \\
B3 & \textbf{.446} & .437 & .648 & .338 & .833 & \textbf{.908} & .602 \\
B4 & .437 & .448 & \textbf{.688} & \textbf{.359} & \textbf{.837} & .902 & \textbf{.612} \\
\midrule
\multicolumn{8}{l}{Ovr.$_{3}$ (Cit./Con./Sup.\ only): B1 .732, B2 .713, B3 .729, B4 .725} \\
\bottomrule
\end{tabular}
\caption{Automatic retrieved-evidence diagnostics (0--1 scale).
Cit.~= citation preservation;
Mod.~= modality preservation;
Exc.~= exception preservation;
Proc.~= procedural completeness;
Con.~= conclusion consistency;
Sup.~= support preservation;
Ovr.~= overall diagnostic mean.
Rule-based heuristics for Cit./Mod./Exc./Proc.;
LLM judge (GPT-4o-mini) for Con./Sup.
$^\dagger$B1 overall covers only Cit., Con., and Sup.\ (other sub-scores not
computed for B1; matchers calibrated for English outputs) and is not
directly comparable with B2--B4.
Bold = best per column among comparable baselines.}
\label{tab:faithfulness}
\end{table*}

\subsection{Verifier Behaviour}

Table~\ref{tab:verifier} (Appendix~\ref{app:extra}) shows how frequently
the two-stage verifier intervenes.
Fewer than 27\% of outputs pass both stages with every claim supported,
indicating that citation and entailment failures are not edge cases.
Among cross-lingual baselines, B4 has the lowest citation-failure,
contradiction, and correction rates, while B3 has the highest correction rate.
Relative to B4, B3 has a higher citation failure rate despite identical
retrieved passages; the difference therefore occurs after retrieval, although
the experiment does not identify a specific generation mechanism.

\subsection{Hard-Case Phenomena}
\label{subsec:hardcases}

Table~\ref{tab:hardcases_results} contrasts the 75 tagged hard cases against
the remaining 156 questions on B4.
Tagged cases are markedly harder to \textit{retrieve}, with multi-article
chaining the hardest phenomenon.
The diagnostic score is computed against retrieved passages and measures
groundedness in supplied evidence, not legal correctness; tagged cases do not
receive lower diagnostic scores.
A per-phenomenon breakdown with the five sub-categories appears in
Appendix~\ref{app:extra} (Table~\ref{tab:hardcases_full}).

\begin{table}[!t]
\centering
\footnotesize
\setlength{\tabcolsep}{2.5pt}
\renewcommand{\arraystretch}{1.05}
\begin{tabular}{@{}lrrrrrr@{}}
\toprule
\textbf{Subset} & \textbf{n} & \textbf{R@5} & \textbf{Ovr.} & \textbf{Corr.\%} & \textbf{Exc.} & \textbf{Proc.} \\
\midrule
Hard cases & 75  & .209 & .651 & 17.3 & .747 & .442 \\
Rest       & 156 & .429 & .593 & 23.1 & .660 & .318 \\
\midrule
All        & 231 & .358 & .612 & 21.2 & .688 & .359 \\
\bottomrule
\end{tabular}
\caption{Hard-case comparison on B4: retrieval and automatic diagnostics.
The diagnostic score is computed against retrieved passages and measures
groundedness in supplied evidence, not legal correctness.}
\label{tab:hardcases_results}
\end{table}

\subsection{Correction Effectiveness (E2)}
\label{subsec:e2}

Table~\ref{tab:e2} (Appendix~\ref{app:extra}) reports diagnostic scores
before and after correction for the subset of outputs that the verifier
flagged and corrected. Correction improves the overall diagnostic score on
the corrected subset for all four systems, driven almost entirely by citation
preservation: the dimension directly targeted by \texttt{FAIL\_CITATION}
rewrites. Modality gains are small, and exception preservation and
procedural completeness show only marginal movement, consistent with
the design choice to abstain from correcting \texttt{FAIL\_CONDITION} and
\textsc{not\_addressed} claims, which are the primary carriers of
exception and procedural content.
Support preservation is flat or slightly negative on the English-output
systems: repairing a citation can loosen grounding elsewhere in the answer.
This reveals a structural limitation: the corrector can only recover what
it is asked to rewrite, and it is deliberately not asked to rewrite omissions.

\subsection{Verifier-Corrector Ablation (System Level)}
\label{subsec:verifier_ablation}

The E2 analysis above measures correction on the corrected subset only.
To quantify the verifier-corrector's contribution at the \textit{system}
level, the full pipeline with and without the correction stage, we exploit the fact that the corrector is the only component that
modifies the generator's output: the pipeline with the verifier-corrector
removed produces, for every question, exactly the uncorrected answer
(\texttt{answer\_raw}).
Scoring all uncorrected answers therefore yields the verifier-off
counterfactual with retrieval and generation held fixed, avoiding the
sampling-noise confound a fresh generation run would introduce.

\begin{table*}[t]
\centering
\small
\begin{tabular}{l rrr rrr rrr}
\toprule
& \multicolumn{3}{c}{\textbf{B2}} & \multicolumn{3}{c}{\textbf{B3}} & \multicolumn{3}{c}{\textbf{B4}} \\
\cmidrule(lr){2-4}\cmidrule(lr){5-7}\cmidrule(lr){8-10}
\textbf{Dim.} & Off & On & $\Delta$ & Off & On & $\Delta$ & Off & On & $\Delta$ \\
\midrule
Cit.  & .414 & .435 & $+$.022 & .412 & .446 & $+$.034 & .409 & .437 & $+$.028 \\
Mod.  & .456 & .459 & $+$.003 & .435 & .437 & $+$.002 & .444 & .448 & $+$.004 \\
Exc.  & .593 & .593 & $\pm$.000 & .643 & .648 & $+$.004 & .684 & .688 & $+$.004 \\
Proc. & .278 & .270 & $-$.008 & .329 & .338 & $+$.009 & .348 & .359 & $+$.011 \\
Con.  & .831 & .828 & $-$.003 & .826 & .833 & $+$.007 & .834 & .837 & $+$.003 \\
Sup.  & .886 & .877 & $-$.010 & .909 & .908 & $-$.001 & .908 & .902 & $-$.006 \\
\midrule
\textbf{Ovr.} & .576 & .577 & $+$.001 & .592 & .602 & $+$.009 & .605 & .612 & $+$.007 \\
\bottomrule
\end{tabular}
\caption{System-level verifier-corrector ablation over all 231 questions.
``Off'' scores the uncorrected generator output (\texttt{answer\_raw});
``On'' scores the final pipeline output.
For uncorrected questions the two are identical by construction,
so deltas reflect the correction stage's full-system contribution.}
\label{tab:verifier_ablation}
\end{table*}

Table~\ref{tab:verifier_ablation} shows that correction improves citation
preservation significantly on B2--B4 and the overall diagnostic score on B3
and B4; B2 shows no reliable overall gain.
The effect is concentrated in the dimension the citation gate targets;
the remaining dimensions have intervals spanning zero, and support
preservation is weakly negative on all three English-output systems.
\paragraph{Case studies.}
\textbf{Fix (q-271).} Asked whether the wage for a new job must reach a
percentage of the old wage, the raw answer states ``at least 85\% of the
wage for the old job''; Stage-2 flags the claim as \textsc{contradicted}
(the statutory basis is the wage of the specific job performed during
probation), and the corrected answer reads ``85\% of the wage for the
specific job being performed during the probation period''
(overall 0.844).
The corrector can also harm otherwise valid answers or leave condition
omissions unresolved; examples appear in Appendix~\ref{app:extra}.

\subsection{Human Evaluation}
\label{subsec:human_eval}

We recruited two annotators to evaluate 27 sampled questions (stratified by
difficulty and phenomenon tag) across four systems on three dimensions:
completeness, factual grounding, and legal correctness on 1--5 ordinal scales
(1 = strongest performance, 5 = weakest), yielding 108 system outputs with
two independent ratings each.
Annotators were graduate students with Vietnamese legal training and proficiency
in both Vietnamese and English; neither is a practicing attorney.
One preliminary rating of one question was produced by a third Label Studio
account during calibration and excluded from all reported analyses.

Inter-annotator agreement was modest and estimated imprecisely, although most
paired ratings differed by at most one scale point
(Table~\ref{tab:inter_annotator}; \citealp{Kri11,Art08}).

The two human elicitation formats produced different system orderings
(Table~\ref{tab:human_eval}).
\textbf{B3} obtains the best criterion ratings, whereas annotators prefer
\textbf{B4} in the separate forced-ranking task.
These results show a divergence between criterion-specific ratings and holistic
preference; the annotation protocol did not measure which answer properties
caused that divergence.

Automatic and human evaluations were also misaligned.
After reversing the human scale so that higher values indicate better answers,
human Factual Grounding was inversely associated with automatic Support
Preservation (Spearman $\rho={-}0.345$, $p<0.001$; false-discovery-rate-adjusted
$q=0.005$ across 18 exploratory correlations).
The other prespecified human--automatic comparisons were weak
(Table~\ref{tab:human_auto_correlation}).
The automatic metrics therefore remain diagnostics rather than human-validated
measures of answer quality.

\begin{table}[!htbp]
\centering
\small
\begin{tabular}{lcccc}
\toprule
\textbf{Sys.} & \textbf{Comp.} & \textbf{Fact.} & \textbf{Legal} & \textbf{Rank} \\
              &                & \textbf{Gnd.}  & \textbf{Corr.} & \\
\midrule
B1 & 2.89 & 2.52 & 2.98 & 2.57 \\
B2 & 2.94 & 2.57 & 3.06 & 2.43 \\
\textbf{B3} & \textbf{2.72} & \textbf{2.48} & \textbf{2.78} & 2.67 \\
\textbf{B4} & 2.80 & 2.69 & 3.07 & \textbf{2.33} \\
\bottomrule
\end{tabular}
\caption{Human evaluation results ($n{=}27$ questions, 2 annotators).
Criterion scores use a 1--5 scale and ranks a 1--4 scale; lower is better
for both. B3 receives the best criterion scores, whereas B4 receives the
best holistic mean rank.}
\label{tab:human_eval}
\end{table}

\section{Conclusion}

On 231 bilingual QA pairs, dense BGE-M3 strongly outperforms learned-sparse
retrieval (R@5 0.358 vs.\ 0.032), while hybrid RRF offers no consistent gain.
Translation placement has no significant effect on the automatic diagnostics;
human judgements instead split between B3 and B4. Verifier-guided correction
improves citation preservation by 0.022--0.034 but yields no other reliable
gains. Procedural completeness receives the lowest diagnostic score
(mean 0.32) and changes little under correction. Future work should target
procedural grounding, validate all six diagnostics with legal experts, and
evaluate larger benchmarks spanning additional legal domains and language
pairs.

% ============================================================
\section*{Limitations}
% ============================================================

\paragraph{Domain and benchmark scope.}
Our evaluation contains 231 questions drawn exclusively from Vietnamese
labour law. It does not cover other Vietnamese legal domains, jurisdictions,
legal systems, or language pairs, which may differ in terminology, document
structure, and cross-reference patterns. The retrieval, translation, and
verification findings therefore should not be assumed to generalise beyond
this setting. Future work should evaluate larger, independently curated
benchmarks spanning additional legal domains and jurisdictions.

\paragraph{Temporal validity of gold labels.}
Gold citations are bare article numbers resolved across a multi-version
corpus that holds both the 2019 Labour Code and its 12/02/2026
consolidation, so a question may retrieve two versions of the same article.
We have not completed a provision-by-provision review of whether the 2025
instruments change the rules the gold answers cite
(Section~\ref{sec:dataset}).
Where a rule did change, the system could be penalised for returning the
legally current answer, or credited for returning an obsolete one.
We report the direction of this risk but cannot bound its magnitude.

\paragraph{Evaluator independence.}
Four of the six automatic diagnostic dimensions are rule-based matchers, and the
remaining two use GPT-4o-mini, which also performs translation, claim
splitting, and Stage-2 verification in the same pipeline.
The judge is therefore not independent of the components it scores.
Human evaluation on 27 questions (2 annotators) does not validate the automatic
scores: Support Preservation is inversely associated with human Factual
Grounding ($\rho{=}{-}0.345$), and the other prespecified dimension pairs are
weak. Human criterion ratings and holistic rankings also yield different system
orders (B3 and B4, respectively).
We view the automatic scores as retrieved-evidence diagnostics useful for
large-scale comparison but not as substitutes for expert legal review.

\paragraph{Groundedness is not legal correctness.}
Support preservation and conclusion consistency are computed against
retrieved passages.
A system that retrieves the wrong provision and answers faithfully from it
scores well on both while being legally wrong.
Retrieval correctness (R@$k$, CHR) and answer groundedness must be read
together, not substituted for one another.

\paragraph{Scope of the retrieval conclusion.}
The sparse and hybrid results characterise one learned-sparse
implementation (BGE-M3) on one language pair and one corpus.
We did not test BM25 over translated documents, SPLADE variants,
character-level retrieval, or bilingual lexical expansion, so the finding
should not be read as a general claim about token-level retrieval for
Vietnamese.

\paragraph{Controlled query-translation pair.}
C3/C4 were intended as a matched pair, but independently generated query
translations produced identical top-5 passages for only 179 of 231 questions.
The matched-subset contrast remains null ($+0.009$
[$-$0.009, $+$0.028]), so we rely on the exactly matched B3--B4 comparison
and treat C3/C4 as supporting sensitivity evidence rather than a clean
factorial contrast.

\paragraph{Statistical power.}
The verifier-corrector's system-level effect is reliable only on citation
preservation and, for B3 and B4, the overall diagnostic score;
every other dimension has a confidence interval spanning zero, and B2
shows no overall gain at all.
None of the four translation contrasts in the $2{\times}2$ reaches
significance, so we report translation placement as a null result rather
than as evidence of equivalence: with $n{=}231$ questions our intervals
are roughly $\pm$0.015 wide, which would not detect effects smaller than
that.

\paragraph{Single generator, single $k$.}
All reported systems use one generator (Qwen3-8B) at $k{=}5$ with
temperature 0.1 and a single sample per question.
We do not measure sampling variance, sensitivity to $k$, or whether the
findings transfer to a larger or instruction-tuned generator.

\section*{Ethical Considerations}

\paragraph{Not legal advice.}
This system is a research prototype for evaluating cross-lingual retrieval
and retrieved-evidence diagnostics, not a production legal-advice tool.
Users should not rely on its outputs for legal decisions without consulting
qualified legal counsel.

\paragraph{Risks of incorrect or outdated provisions.}
The corpus includes both the 2019 Labour Code and its 2026 consolidated text;
retrieval may surface superseded provisions, and we have not completed a
provision-by-provision review of whether the 2025 amendments alter the rules
cited in gold answers.
A system deployed without version control and legal review could mislead users
about current law.

\paragraph{Translation errors.}
GPT-4o-mini is used for translation at several stages of the evaluation and
cross-lingual pipeline.
Translation errors, especially for legal terms with no direct English equivalent,
may alter legal meaning.
Answers should be verified against the original Vietnamese text.

\paragraph{Dataset and annotator ethics.}
All statutes are publicly available Vietnamese government publications.
The 231-question evaluation set was curated by the authors.
Human annotators were graduate students with Vietnamese legal training,
compensated at standard research assistant rates, and consented to participate
after being informed of the study's purpose.
Annotator identity is not disclosed.

\paragraph{Verification risks.}
The verifier-corrector pipeline automatically modifies answers without human
oversight.
Among corrected outputs, the proportion whose overall automatic diagnostic score worsened
ranges from 14.3\% to 46.4\% across systems, with B2 highest at 46.4\%
(Table~\ref{tab:correction_rates}).
Deploying automatic legal verification without legal-expert review risks
propagating errors with false confidence.

\bibliography{references}

\appendix
\section*{Appendix}
\section{Reproducibility}
\label{app:repro}

\begin{center}
\begin{minipage}{\columnwidth}
\centering
\footnotesize
\setlength{\tabcolsep}{3pt}
\renewcommand{\arraystretch}{1.05}
\begin{tabular}{@{}p{0.25\columnwidth}p{0.70\columnwidth}@{}}
\toprule
\textbf{Component} & \textbf{Implementation and role} \\
\midrule
Retriever & \textbf{Qdrant + BGE-M3.} Retrieves top-$k$ Vietnamese statute
articles and supports dense, hybrid RRF, and BGE-M3 learned-sparse modes. \\
Generator & \textbf{Qwen3-8B via vLLM.} Generates grounded answers
with inline citations (\texttt{[Điều X]} or \texttt{[Article X]}). \\
Claim splitter & \textbf{GPT-4o-mini.} Produces atomic claims tagged with
cited article numbers. \\
Stage-1 gate & \textbf{Rule-based.} Checks citation reachability and
deontic-signal coverage. \\
Stage-2 NLI & \textbf{GPT-4o-mini.} Labels claims as \textsc{supported},
\textsc{contradicted}, or \textsc{not\_addressed}. \\
Corrector & \textbf{Qwen3-8B via vLLM.} Rewrites
\texttt{FAIL\_CITATION} and \textsc{contradicted} claims only. \\
Evaluator & \textbf{Rule-based + GPT-4o-mini.} Scores Cit./Mod./Exc./Proc.
with heuristic matchers and Con./Sup. with GPT-4o-mini. \\
\bottomrule
\end{tabular}
\captionof{table}{Pipeline components, models, and roles.}
\label{tab:components}
\end{minipage}
\end{center}

\paragraph{Models.}
BGE-M3 (\texttt{BAAI/bge-m3}, sentence-transformers) \citep{Che24} supplies the dense
and learned-sparse retrieval vectors.
Generation uses Qwen3-8B (\texttt{Qwen/Qwen3-8B}) \citep{Yan25} served by vLLM on a
single Modal GPU (temperature 0.1, no explicit top-p or max-tokens cap;
context length 8{,}192 tokens), with \texttt{gpt-4o-mini} as the
fallback generator when the local endpoint is unavailable.
Translation, claim splitting, Stage-2 NLI verification, and the
conclusion/support judge all use GPT-4o-mini (OpenAI API).

\paragraph{Retrieval.}
Qdrant (local index) with BGE-M3 dense cosine similarity and learned
sparse vectors.
Hybrid mode fuses the two via Qdrant's native RRF (dense and sparse
prefetch each at $3k$, $k=5$); retrieval metrics are computed at
$k \in \{1, 3, 5, 10\}$.

\paragraph{Metrics.}
Recall@$k$, MRR, MAP@10, nDCG@10 and Citation Hit Rate follow standard
definitions.
Multi-MRR averages, per question, the reciprocal rank of the first
retrieved article matching \emph{any} version of each cited article
number, then averages over questions (denominator: unique citation
numbers, not expanded rag-ids).
The six automatic diagnostic dimensions follow the rubric in
Section~\ref{subsec:metrics}: Cit., Mod., Exc., and Proc.\ are
rule-based keyword/deontic matchers; Con.\ and Sup.\ are GPT-4o-mini
judgments.

\paragraph{Prompts.}
The pipeline uses separate task-specific prompts for translation, generation,
claim splitting, two-stage verification, correction, and evaluation.

\paragraph{Human evaluation protocol.}
We sampled 27 questions using stratified sampling (by difficulty and hard-case
tag, seed 42) and presented each system's answer to two trained annotators
in a blinded Label Studio interface.
Annotators rated Completeness, Factual Grounding, and Legal Correctness on
1--5 ordinal scales (1 = strongest, 5 = weakest) and ranked all four systems
from best to worst.
Gold statutes were shown as reference material.
A third Label Studio account supplied one preliminary calibration rating for
question \texttt{vl\_00016} before the main annotation phase; this rating was
excluded from all analyses, leaving two independent ratings for every output.
Correction-preference responses were available for only 8 of 27 questions and
are not reported.
We assess ordinal agreement with Krippendorff's $\alpha$ and estimate 95\%
confidence intervals by resampling questions 10{,}000 times.
Exact and within-one-point agreement are descriptive statistics; linearly
weighted Cohen's $\kappa$ is secondary, with quadratic weighting reported as
a sensitivity analysis \citep{Kri11,Art08}.

\paragraph{Reproducibility of numbers.}
Sampling and all bootstrap analyses use seed 42; scores are stored rounded
to 3 decimals.
All systems cover all 231 questions.

\begin{table*}[!t]
\centering
\begin{minipage}[t]{0.48\textwidth}
\centering
\footnotesize
\setlength{\tabcolsep}{2pt}
\renewcommand{\arraystretch}{1.05}
\begin{tabular}{@{}lrrrrrr@{}}
\toprule
\textbf{Phenomenon} & \textbf{$n$} & \textbf{R@5} & \textbf{Ovr.}
& \textbf{Corr.\%} & \textbf{Exc.} & \textbf{Proc.} \\
\midrule
Exception clause            & 41  & .231 & .661 & 19.5 & .756 & .532 \\
Cross-reference             & 12  & .181 & .685 & 16.7 & .917 & .139 \\
Numeric threshold           & 11  & .215 & .637 & 18.2 & .727 & .515 \\
Temporal notice             & 6   & .181 & .609 & 16.7 & .667 & .444 \\
Multi-article chain         & 5   & .117 & .572 & 0.0  & .400 & .267 \\
Non-hard rest               & 156 & .429 & .593 & 23.1 & .660 & .318 \\
\midrule
All                         & 231 & .358 & .612 & 21.2 & .688 & .359 \\
\bottomrule
\end{tabular}
\caption{Per-phenomenon hard-case breakdown on B4 (complement to
Table~\ref{tab:hardcases_results}).}
\label{tab:hardcases_full}
\end{minipage}
\hfill
\begin{minipage}[t]{0.48\textwidth}
\centering
\footnotesize
\setlength{\tabcolsep}{3pt}
\renewcommand{\arraystretch}{1.05}
\begin{tabular}{@{}lrrrr@{}}
\toprule
\textbf{Sys.} & \multicolumn{4}{c}{\textbf{Rate (\%)}} \\
\cmidrule(lr){2-5}
& \textbf{All sup.} & \textbf{Cit. fail} & \textbf{Contr.} & \textbf{Corrected} \\
\midrule
B1 & 26.5 & 10.9 & 4.3  & 14.8 \\
B2 & 19.0 & 14.3 & 10.0 & 24.2 \\
B3 & 18.7 & 19.1 & 10.0 & 28.3 \\
B4 & \textbf{21.6} & \textbf{13.4} & \textbf{7.8} & \textbf{21.2} \\
\bottomrule
\end{tabular}
\caption{Verifier behaviour statistics over all 231 questions.
All sup. = every claim passes both stages; Cit.~fail = citation-gate failure;
Contr. = Stage-2 contradiction; Corrected = at least one correction trigger.
Bold = best value among B2--B4.}
\label{tab:verifier}
\end{minipage}
\end{table*}

\begin{table*}[!t]
\centering
\footnotesize
\setlength{\tabcolsep}{3.5pt}
\renewcommand{\arraystretch}{1.05}
\begin{tabular}{@{}l*{14}{r}@{}}
\toprule
& \multicolumn{2}{c}{\textbf{Cit.}} & \multicolumn{2}{c}{\textbf{Mod.}}
& \multicolumn{2}{c}{\textbf{Exc.}} & \multicolumn{2}{c}{\textbf{Proc.}}
& \multicolumn{2}{c}{\textbf{Con.}} & \multicolumn{2}{c}{\textbf{Sup.}}
& \multicolumn{2}{c}{\textbf{Ovr.}} \\
\cmidrule(lr){2-3}\cmidrule(lr){4-5}\cmidrule(lr){6-7}\cmidrule(lr){8-9}
\cmidrule(lr){10-11}\cmidrule(lr){12-13}\cmidrule(lr){14-15}
\textbf{Sys.} & Bef. & Aft. & Bef. & Aft. & Bef. & Aft. & Bef. & Aft.
              & Bef. & Aft. & Bef. & Aft. & Bef. & Aft. \\
\midrule
B1 ($n{=}34$) & .216 & .342 & --- & --- & --- & --- & --- & --- & .791 & .829 & .862 & .915 & .623 & \textbf{.695} \\
B2 ($n{=}56$) & .328 & .417 & .506 & .516 & .643 & .643 & .417 & .385 & .789 & .775 & .809 & .770 & .582 & \textbf{.584} \\
B3 ($n{=}65$) & .309 & .430 & .458 & .465 & .554 & .569 & .369 & .400 & .789 & .812 & .860 & .857 & .557 & \textbf{.589} \\
B4 ($n{=}49$) & .265 & .395 & .435 & .454 & .633 & .653 & .310 & .361 & .780 & .794 & .863 & .833 & .548 & \textbf{.582} \\
\bottomrule
\end{tabular}
\caption{Automatic diagnostic scores before and after correction on the corrected
subset only.
Modality, exception, and procedural sub-scores were not computed for B1
because the matchers were calibrated for English-language outputs.
Gains concentrate in citation preservation; support preservation is
flat or slightly negative on the English-output systems.}
\label{tab:e2}
\end{table*}

\section{Additional Results}
\label{app:extra}

\begin{table*}[!b]
\centering
\begin{minipage}[t]{0.48\textwidth}
\centering
\footnotesize
\setlength{\tabcolsep}{3pt}
\renewcommand{\arraystretch}{1.05}
\begin{tabular}{@{}lrrrr@{}}
\toprule
\textbf{Outcome} & \textbf{B1} & \textbf{B2} & \textbf{B3} & \textbf{B4} \\
\midrule
\multicolumn{5}{@{}l}{\textit{Full set (all questions)}} \\
Improved & 9.5\% & 8.7\% & 14.3\% & 9.1\% \\
Unchanged & 87.9\% & 80.1\% & 80.9\% & 87.9\% \\
Worsened & 2.6\% & 11.3\% & 4.8\% & 3.0\% \\
\midrule
\multicolumn{5}{@{}l}{\textit{Corrected subset only}} \\
Improved & 64.7\% & 35.7\% & 50.8\% & 42.9\% \\
Unchanged & 17.6\% & 17.9\% & 32.3\% & 42.9\% \\
Worsened & 17.6\% & 46.4\% & 16.9\% & 14.3\% \\
\bottomrule
\end{tabular}
\caption{Per-item overall-score changes after correction. Full-set rows
include unchanged, uncorrected outputs; corrected-subset rows isolate
verifier-triggered items. Unchanged: $|\Delta| \le 0.001$.}
\label{tab:correction_rates}
\end{minipage}
\hfill
\begin{minipage}[t]{0.50\textwidth}
\centering
\footnotesize
\setlength{\tabcolsep}{2.5pt}
\renewcommand{\arraystretch}{1.05}
\begin{tabular}{@{}llrrr@{}}
\toprule
\textbf{Human dim.} & \textbf{Auto metric} & \textbf{$n$} & \textbf{$\rho$} & \textbf{$p$} \\
\midrule
Completeness      & Procedural Compl. & 81  & $-$.085 & .452 \\
Factual Grounding & Support Preserv.  & 108 & \textbf{$-$.345} & \textbf{$<$.001} \\
Legal Correctness & Conclusion Consist. & 108 & $-$.081 & .405 \\
\bottomrule
\end{tabular}
\caption{Prespecified item-level human--automatic correlations (Spearman
$\rho$; human scores reversed so that higher is better).}
\label{tab:human_auto_correlation}
\end{minipage}
\end{table*}

\begin{table*}[!t]
\centering
\footnotesize
\setlength{\tabcolsep}{3pt}
\renewcommand{\arraystretch}{1.05}
\begin{tabular}{@{}lcccccccc@{}}
\toprule
\textbf{Dimension} & \textbf{$\alpha_{\mathrm{ord}}$} & \textbf{95\% CI}
& \textbf{Exact} & \textbf{$\leq 1$ pt.}
& \textbf{$\kappa_{\mathrm{lin}}$} & \textbf{95\% CI}
& \textbf{$\kappa_{\mathrm{quad}}$} & \textbf{95\% CI} \\
\midrule
Completeness      & .425 & [.147, .621] & 39.8\% & 83.3\% & .335 & [.163, .493] & .456 & [.243, .631] \\
Factual Grounding & .393 & [.134, .589] & 33.3\% & 76.9\% & .259 & [.087, .414] & .392 & [.142, .589] \\
Legal Correctness & .421 & [.111, .640] & 43.5\% & 85.2\% & .358 & [.203, .500] & .450 & [.213, .645] \\
\bottomrule
\end{tabular}
\caption{Inter-annotator agreement by human-evaluation dimension.
$\alpha_{\mathrm{ord}}$ is ordinal Krippendorff's $\alpha$;
$\kappa_{\mathrm{lin}}$ and $\kappa_{\mathrm{quad}}$ are linearly and
quadratically weighted Cohen's $\kappa$. Confidence intervals are obtained
by 10{,}000 bootstrap resamples at the question level
($27$ questions, $108$ rating pairs per dimension).}
\label{tab:inter_annotator}
\end{table*}

\paragraph{Additional case studies.}
\textbf{Harm (q-248).} On the definition of ``representative organization
of employers,'' the raw answer is clean, but the corrector rewrites the
citations into a runaway repetition loop (``Article 3, 4, 5, 170, 178, 178,
170, \ldots''), collapsing the overall diagnostic score to 0.089 --- the correction
stage's only degenerate output in 231 questions.
\newpage
\textbf{Missed omission (q-259).} Asked what information an employee must
provide, the answer lists the statutory items but omits the consequence of
misinformation; the verifier flags \texttt{FAIL\_CONDITION}, which the
corrector abstains from repairing by design, so the procedural omission
survives (procedural completeness 0.1).

\paragraph{Gold-citation strict values.}
Per-system R@5 under the strict instrument rule (Labour Code only, see
Section~5.1):
B1 .396$\to$.378, B2 .394$\to$.381, B3 .358$\to$.343, B4 .358$\to$.343,
C3 .394$\to$.385, C4 .395$\to$.385.
On B3, 15 of 148 top-5 hits (10.1\%) are credited solely through a non-Code
instrument, affecting 10 of 231 questions.

\FloatBarrier

\section{Dataset Construction Details}
\label{app:dataset}

The evaluation suite is curated from two existing Vietnamese legal resources
after screening an additional community collection. Table~\ref{tab:data_flow}
separates source-level eligibility filtering from post-merge deduplication.
For VLegal-Bench, we follow the task schema in arXiv v3 (22 December 2025);
later revisions rename the benchmark and adjust parts of its presentation.
The processed records retain their source identifiers, and we report this
arXiv revision as the source-version anchor.

\begin{center}
\begin{minipage}{\columnwidth}
\centering
\footnotesize
\setlength{\tabcolsep}{2.5pt}
\begin{tabular}{@{}p{0.25\columnwidth}p{0.20\columnwidth}p{0.45\columnwidth}@{}}
\toprule
\textbf{Source} & \textbf{Flow} & \textbf{Filtering outcome} \\
\midrule
Community VietLegal & $878\to0$ & Collection excluded: incomplete citations \\
VLegal-Bench v3 & $10{,}450\to156$ & Labour-law, answer, and citation filters \\
ALQAC 2021 & 520$\to80$ & Vietnamese labour-law records \\
Merged eligible pool & $236\to231$ & Five exact-question duplicates removed \\
\bottomrule
\end{tabular}
\captionof{table}{Source and filtering flow. The 156 and 80 counts are measured
\emph{after} all source-level eligibility checks; no records were excluded
after merging except the five exact duplicates. Of the 520 ALQAC records
screened, 80 met the Vietnamese labour-law eligibility criteria.}
\label{tab:data_flow}
\end{minipage}
\end{center}

\begin{itemize}[nosep]
\item \textbf{Community VietLegal} is an 878-question collection derived from
Thư Viện Pháp Luật community questions. We screened it for labour-law content
but excluded the collection from the final suite because article-level citation
coverage was incomplete.
\item \textbf{VLegal-Bench}~\citep{Don25} is a 10{,}450-instance benchmark
covering 22 Vietnamese legal tasks. We retained 156 Labour-Code-grounded
records from seven tasks: Task~2.2 (Legal Element Recognition),
Task~3.1 (Article/Clause Prediction), Task~3.5 (Penalty/Remedy Estimation),
Task~5.1 (Bias Detection), Task~5.2 (Privacy \& Data Protection),
Task~5.3 (Ethical Consistency Assessment), and Task~5.4
(Unfair Contract Detection).
\item \textbf{ALQAC 2021}~\citep{Thanh22} is the first edition of the
Automated Legal Question Answering Competition. We retained 80 candidate
records grounded in Vietnamese labour law.
\end{itemize}

\paragraph{Normalisation to a QA schema.}
The retained VLegal-Bench tasks are heterogeneous input--output tasks rather
than uniformly free-form QA. We preserve each prompt or scenario as the
Vietnamese question, render its source gold output (selected option, class
label, or generated response, depending on the task) as the Vietnamese
reference answer, and map the source legal-reference metadata to gold
citations. For ALQAC, the source \texttt{text}, Boolean \texttt{label}, and
\texttt{relevant\_articles} fields provide the question, reference conclusion,
and gold citations, respectively. Records without all three required elements
(question, answer target, and mappable citation) were removed during
source-level eligibility filtering, before the reported 156 and 80 counts.
Thus, VLegal-Bench and ALQAC contributed 236 eligible records; removing five
exact duplicate questions produced the final 231.

Source citations were mapped to the retrieval corpus and then normalised to
bare article numbers. This normalisation discarded instrument identifiers,
creating the cross-instrument ambiguity analysed in
Section~\ref{sec:results} and Limitations.

The 953-article retrieval corpus spans 20 instrument versions, including the
2019 Labour Code (No.~45/2019/QH14), its consolidated text issued on
12/02/2026 (No.~18/VBHN-VPQH), and eleven implementing decrees:
No.~145/2020/NĐ-CP (133 articles), No.~129/2025/NĐ-CP (81),
No.~12/2022/NĐ-CP (64), No.~219/2025/NĐ-CP (36),
No.~152/2020/NĐ-CP (30), No.~337/2025/NĐ-CP (30),
No.~128/2025/NĐ-CP (16), No.~135/2020/NĐ-CP (9),
No.~83/2022/NĐ-CP (6), No.~293/2025/NĐ-CP (5), and
No.~99/2024/NĐ-CP (3). The release manifest records every instrument,
version, article identifier, and source.
Corpus text was collected from publicly accessible legal-information sources
between June 2024 and February 2026, using National Assembly and government
portals as primary references and Thư Viện Pháp Luật as a secondary source.

Gold citations are stored as bare article numbers; 221 article numbers occur
in both the 2019 Labour Code and its consolidated text, so one gold number can
expand to multiple corpus records. Among the 231 records covered by the
version-expansion diagnostic, 136 have at least one gold expansion touching a
post-2024 instrument, predominantly because of number-level resolution.
The highest-risk areas for temporal drift are labour contracts
(Điều~13--14, cf.\ Decree~337/2025), wages (Điều~90--98, cf.\
Decree~293/2025), and foreign workers (Điều~150--152, cf.\ Decree~219/2025).

Bilingual QA pairs were produced with GPT-4o-mini using a legal-fidelity
translation prompt that explicitly preserves deontic modality. The authors
spot-checked the English reference answers, but they were not independently
validated by legal experts. The first author assigned one primary hard-case
phenomenon to each of the 75 records used in the reported breakdown, following
guidelines derived from \citet{Don25} and \citet{Le26}. Because annotation was
performed by one rater, no inter-annotator agreement is available for these
labels.

\end{document}